\documentclass[sigconf]{acmart}
\usepackage{multirow}
\usepackage{algpseudocode}
\usepackage{makecell}
\usepackage[linesnumbered,ruled,vlined]{algorithm2e}
\SetKwInput{KwInput}{Input}    % Set the Input
\SetKwInput{KwOutput}{Output}  % Set the Output
\AtBeginDocument{%
  }

\setcopyright{acmlicensed}
\copyrightyear{2018}
\acmYear{2018}
\acmDOI{XXXXXXX.XXXXXXX}

\acmConference[Conference acronym 'XX]{Make sure to enter the correct
  conference title from your rights confirmation emai}{June 03--05,
  2018}{Woodstock, NY}
\acmISBN{978-1-4503-XXXX-X/18/06}

\begin{document}

\title{An Efficient and Generalizable Symbolic Regression Method for Time Series Analysis}

\author{Yi Xie}
\affiliation{%
  \institution{Shanghai Key Lab of Data Science, School of Computer Science, Fudan University}
  \city{Shanghai}
  \country{China}}
\email{yixie18@fudan.edu.cn}

\author{Tianyu Qiu}
\affiliation{%
  \institution{Shanghai Key Lab of Data Science, School of Computer Science, Fudan University}
  \city{Shanghai}
  \country{China}}
\email{tyqiu22@fudan.edu.cn}

\author{Yun Xiong}
\affiliation{%
  \institution{Shanghai Key Lab of Data Science, School of Computer Science, Fudan University}
  \city{Shanghai}
  \country{China}}
\email{yunx@fudan.edu.cn}

% \author{Hongrun Ren}
% \affiliation{%
%   \institution{Shanghai Key Lab of Data Science, School of Computer Science, Fudan University}
%   \city{Shanghai}
% \email{renhr20@fudan.edu.cn}

% \author{Shuaibin Huang}
% \affiliation{%
%   \institution{Shanghai Key Lab of Data Science, School of Computer Science, Fudan University}
%   \city{Shanghai}
% \email{sbhuang22@fudan.edu.cn}

\author{Xiuqi Huang}
\affiliation{%
  \institution{MoE Key Lab of Artificial Intelligence, Shanghai Jiao Tong University}
  \city{Shanghai}
  \country{China}}
\email{huangxiuqi@sjtu.edu.cn}

\author{Xiaofeng Gao}
\affiliation{%
  \institution{MoE Key Lab of Artificial Intelligence, Shanghai Jiao Tong University}
  \city{Shanghai}
  \country{China}}
\email{gao-xf@cs.sjtu.edu.cn}

\author{Chao Chen}
\affiliation{%
  \institution{College of Computer Science, Chongqing University}
  \city{Chongqing}
  \country{China}}
\email{cschaochen@cqu.edu.cn}

\renewcommand{\shortauthors}{Trovato et al.}

\begin{abstract}

  Time series analysis and prediction methods currently excel in quantitative analysis, 
  offering accurate future predictions and diverse statistical indicators, 
  but generally falling short in elucidating the underlying evolution patterns 
  of time series. To gain a more comprehensive understanding and provide insightful 
  explanations, we utilize symbolic regression techniques to derive explicit 
  expressions for the non-linear dynamics in the evolution of time series variables.
  However, these techniques face challenges in computational efficiency and 
  generalizability across diverse real-world time series data. To overcome these 
  challenges, we propose \textbf{N}eural-\textbf{E}nhanced \textbf{Mo}nte-Carlo 
  \textbf{T}ree \textbf{S}earch (NEMoTS) for time series. 
  NEMoTS leverages the exploration-exploitation balance of Monte-Carlo Tree Search (MCTS), 
  significantly reducing the search space in symbolic regression and improving 
  expression quality.
  Furthermore, by integrating neural networks with MCTS, NEMoTS not only capitalizes 
  on their superior fitting capabilities to concentrate on more pertinent operations 
  post-search space reduction, but also replaces the complex and time-consuming 
  simulation process, thereby substantially improving computational efficiency and 
  generalizability in time series analysis. NEMoTS offers an efficient and comprehensive approach to 
  time series analysis. Experiments with three real-world datasets demonstrate 
  NEMoTS's significant superiority in performance, efficiency, reliability, and interpretability, 
  making it well-suited for large-scale real-world time series data.

\end{abstract}

\begin{CCSXML}
<ccs2012>
   <concept>
       <concept_id>10010147.10010178.10010205.10010210</concept_id>
       <concept_desc>Computing methodologies~Game tree search</concept_desc>
       <concept_significance>300</concept_significance>
       </concept>
   <concept>
       <concept_id>10010147.10010148.10010149.10010154</concept_id>
       <concept_desc>Computing methodologies~Hybrid symbolic-numeric methods</concept_desc>
       <concept_significance>500</concept_significance>
       </concept>
   <concept>
       <concept_id>10002950.10003648.10003688.10003693</concept_id>
       <concept_desc>Mathematics of computing~Time series analysis</concept_desc>
       <concept_significance>500</concept_significance>
       </concept>
   <concept>
       <concept_id>10010147.10010257.10010293.10010294</concept_id>
       <concept_desc>Computing methodologies~Neural networks</concept_desc>
       <concept_significance>300</concept_significance>
       </concept>
   <concept>
       <concept_id>10002951.10003227.10003351.10003446</concept_id>
       <concept_desc>Information systems~Data stream mining</concept_desc>
       <concept_significance>500</concept_significance>
       </concept>
 </ccs2012>
\end{CCSXML}

\ccsdesc[300]{Computing methodologies~Game tree search}
\ccsdesc[500]{Computing methodologies~Hybrid symbolic-numeric methods}
\ccsdesc[500]{Mathematics of computing~Time series analysis}
\ccsdesc[300]{Computing methodologies~Neural networks}
\ccsdesc[500]{Information systems~Data stream mining}

\keywords{Symbolic Regression, Analytical Expression, Neural-Enhanced Monte-Carlo Tree Search, Time Series, Efficiency, Generalizability }

\received{20 February 2007}
\received[revised]{12 March 2009}
\received[accepted]{5 June 2009}

\maketitle

\section{Introduction}
\begin{figure}
  \centering
  \includegraphics[width=\linewidth]{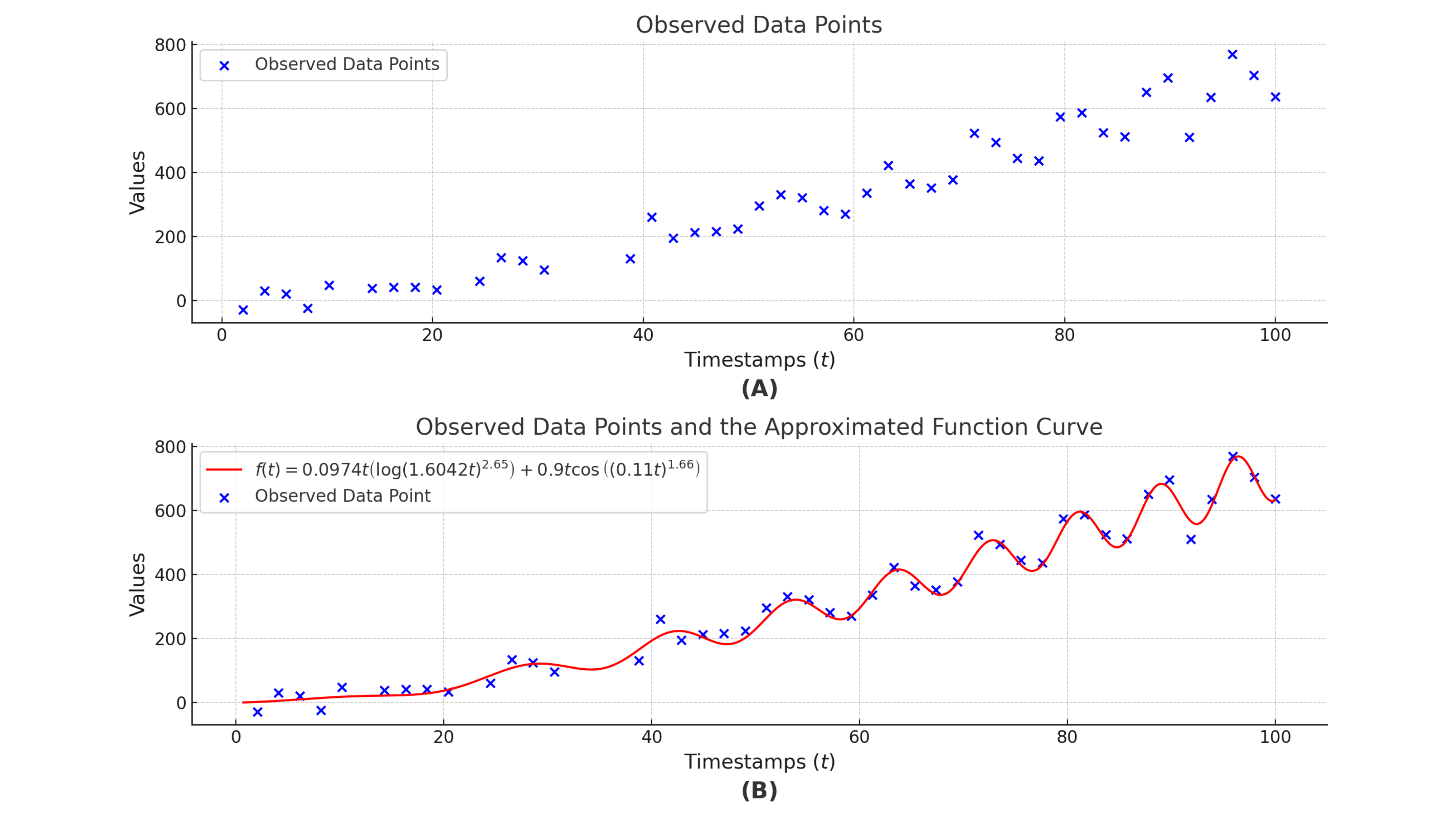}
  \caption{(A) Blue crosses represent the observed values; and (B)
  red curve represents the fitted analytical expression of the time series 
  data.
  }\label{Example}
\end{figure}

Popular time series modeling and analysis frameworks, such as Auto-Regressive Integrated Moving Average (ARIMA) \cite{RW_arima,RW_arima2}, Kalman Filter \cite{Kalman}, Gradient Boosting Decision Tree (GBDT) \cite{GBDT1,GBDT2}, Recurrent Neural Networks (RNN) \cite{RNN1,RNN2,RNN3}, Temporal Convolutional Networks (TCN) \cite{TCN1,TCN2}, and self-attention based methods (X-formers) \cite{Transformer,Informer,Autoformer,TransformerTSSurvey}, are proficient in quantitative analysis. Yet, they often struggle to reveal the underlying patterns in time series data. These approaches mainly concentrate on \textbf{how} data changes over time, but typically neglect \textbf{what} instigates these changes and \textbf{why} specific patterns arise. An analytical expression of time series over time can
identify global evolution patterns in various timeframes \cite{Symb_intro1,Symb_intro2,Symb_Survey}. For example, Fig. \ref{Example} (A) offers limited insights, such as increasing values and periodic oscillations with growing amplitude, due to the lack of an explicit expression. Conversely, Fig. \ref{Example} (B) demonstrates the expression $f(t) = 0.0974t\left(\log(1.6042t)^{2.65}\right) + 0.9t \cos\left((0.11t)^{1.66}\right)$. This reveals a logarithmic trend and a seasonal component with significant cyclical fluctuations, driven by the cosine function. As $t$ increases, the amplitude and frequency of these fluctuations also increase linearly, offering more profound insights than surface-level observations.

Symbolic regression, a classical and highly interpretable machine learning approach, effectively connects inputs and outputs using mathematical expressions made of basic functions, as highlighted in \cite{Symb_Survey,Symb_intro2}. Particularly adept at approximating time-varying physical systems, symbolic regression uses explicit analytical expressions in a data-driven way to skillfully reveal nonlinear system dynamics without prior constraints
(\emph{e.g.}, linear assumptions, polynomial assumptions, or trigonometric function assumptions)  \cite{Symb_dynamics}. In time series analysis, this technique not only provides qualitative insights but also enables an in-depth quantitative examination of fundamental evolutionary processes \cite{Symb_intro1,Symb_intro2}. Unlike traditional quantitative methods, symbolic regression delves deeper into the intrinsic dynamics of time series itself, offering substantial insights into the \textbf{what} and \textbf{why} behind evolution. This method excels in analyzing complex, nonlinear systems where standard modeling techniques might fail to grasp system intricacies. Therefore, symbolic regression introduces a new perspective in understanding and interpreting complex time series systems, greatly enhancing our knowledge of their dynamics in both depth and breadth.

However, current symbolic regression techniques, mainly designed for fitting specific sample, and based on combinatorial optimization methods, often depend on complex heuristic designs to fit a particular case. They use simulation or search algorithms to generate expressions matching that case \cite{Symb_intro1,Symb_intro2}. These methods face challenges such as computational inefficiency, high complexity, and restricted generalization abilities, especially when handling larger datasets \cite{Symb_SurveySearchSpace,Symb_SurveyGen,Symb_Effi}. The increased computational requirements and extended model search durations, particularly in extensive iterative processes, diminish their effectiveness in big data scenarios. Additionally, because these techniques concentrate on fitting specific samples, they struggle to identify common patterns across different samples and lack broader learning capabilities. This not only hinders performance improvement but also underutilizes the rich knowledge embedded in extensive datasets

To address the limitations of current symbolic regression methods, we introduce the \textbf{N}eural-\textbf{E}nhanced \textbf{Mo}nte-Carlo \textbf{T}ree \textbf{S}earch (NEMoTS) for time series analysis. NEMoTS utilizes the Monte-Carlo Tree Search (MCTS) framework, where expressions are represented as tree structures. This design is consistent with the structural integrity of parse trees and adheres to context-free grammar rules, guaranteeing the validity of the generated expressions \cite{Symb_ExpTree1,Symb_ExpTree2}. By leveraging the inherent balance between exploration and exploitation in MCTS, NEMoTS significantly narrows the search space for specific samples, enhancing search efficiency and the quality of expressions relative to other methods \cite{Symb_SPL,Symb_MCTS}.
To tackle issues like the exponential growth of the search space, high computational demands, and limited generalization, NEMoTS incorporates neural networks into the MCTS framework. This fusion utilizes the strong learning capabilities of neural networks to enhance MCTS's performance. During the selection phase, neural network predictions guide the selection of high-potential nodes, focusing on the search. In the simulation phase, neural networks replace complicated fast random simulations, utilizing their advanced fitting capabilities for state assessment, thereby streamlining MCTS operations and boosting efficiency. The integration of neural networks also enables NEMoTS to learn from extensive datasets, augmenting both the model's fitting accuracy and its generalization capacity.

NEMoTS consists of four primary components: a pre-defined basic function library, Monte-Carlo Tree Search (MCTS), a policy-value network, and a coefficient optimizer. At its core, MCTS guides the process. In MCTS, the selection and simulation phases are influenced by the policy-value network's assessment and output regarding the overall state of the expression. Each operation within MCTS and the resulting expression originate from a pre-defined basic function library. MCTS produces an initial expression 'backbone', which lacks numerical coefficients. These are then refined by the coefficient optimizer to create a full expression.
Expanding on ideas from \cite{Symb_SPL}, we integrate a Symbolic Augmentation Strategy (SAS) during training. SAS improves the simulation phase of the Monte-Carlo tree search by accumulating high-rewarded composite functions. This approach is akin to frequent pattern mining \cite{Symb_FreqItem}, involving the random amalgamation of various basic functions to identify frequent, high-rewarded composite function patterns. These frequently occurring composite functions are then added to the function library based on their average rewards, significantly enhancing the model’s fitting abilities.

We carried out comprehensive experiments on three real-world time series datasets. The outcomes reveal that NEMoTS not only excels in symbolic regression tasks for time series, exhibiting exceptional fitting ability and efficiency, but also demonstrates superior performance in extrapolation with the expressions it derives, which implies the reliability of the expressions.

The key contributions of this paper are outlined as follows:
\begin{itemize}
\item \textbf{Application of Symbolic Regression in Time Series Analysis:} We utilize symbolic regression to enhance the analysis and understanding of time series data, especially in qualitative aspect. The integration of Monte-Carlo Tree Search (MCTS) in symbolic regression for time series leads to the discovery of high-quality, valid expressions, providing new insights into time series analysis.
\item \textbf{Advancing Symbolic Regression with Neural Networks through MCTS:} To overcome the inefficiencies and generalization limits of traditional MCTS in larger-scale time series data, neural networks have been incorporated into the framework. This advancement not only increases the model's efficiency but also expands its learning and generalization capabilities, resulting in enhanced performance.
\item \textbf{Development of the NEMoTS Model:} Building upon these innovations, we present the Neural-Enhanced Monte-Carlo Tree Search for Time Series (NEMoTS), specifically tailored for symbolic regression in time series. The unique inclusion of a symbolic augmentation strategy, inspired by frequent pattern mining, further boosts the model's performance.
\item \textbf{Extensive Experimental Validation:} Through comprehensive experiments on three real-world time series datasets, NEMoTS demonstrates its remarkable capability and efficiency in symbolic regression tasks for time series.
\end{itemize}

\section{Problem Definition}
We formally define our task, similar to the classical Empirical Risk Minimization (ERM) approach \cite{Symb_ERM}.

\noindent\textbf{Input:} Given a time series $\mathcal{D}={(t_{i},v_{i})}_{i=0}^{N-1}$ 
containing $N$ records, where $t_{i}\in{\mathbb{R}}$ represents the timestamp 
and $v_{i}\in{\mathbb{R}}$ represents the value corresponding to the timestamp $t_{i}$.

\noindent\textbf{Objective:} The goal is to discover an analytical expression 
$f(\cdot)$ and evaluate it using the following reward function:
\begin{equation}
\mathcal{R}=\frac{\eta^{s}}{1+\sum_{i=0}^{N-1}\sqrt{(v_{i}-f(t_{i}))^{2}}} , \label{reward}
\end{equation}
where $\eta$ is a constant slightly less than $1$, and $s$ denotes the size of the 
generated analytical expression. Generally, the value of this reward function ranges 
between 0 and 1, balancing the complexity of the generated expression and its fitting 
degree. The closer it is to 1, the simpler the discovered expression and the higher 
the achieved fitting accuracy.

\section{Neural-Enhanced MCTS}

\begin{figure*}
  \centering
  \includegraphics[width=\linewidth]{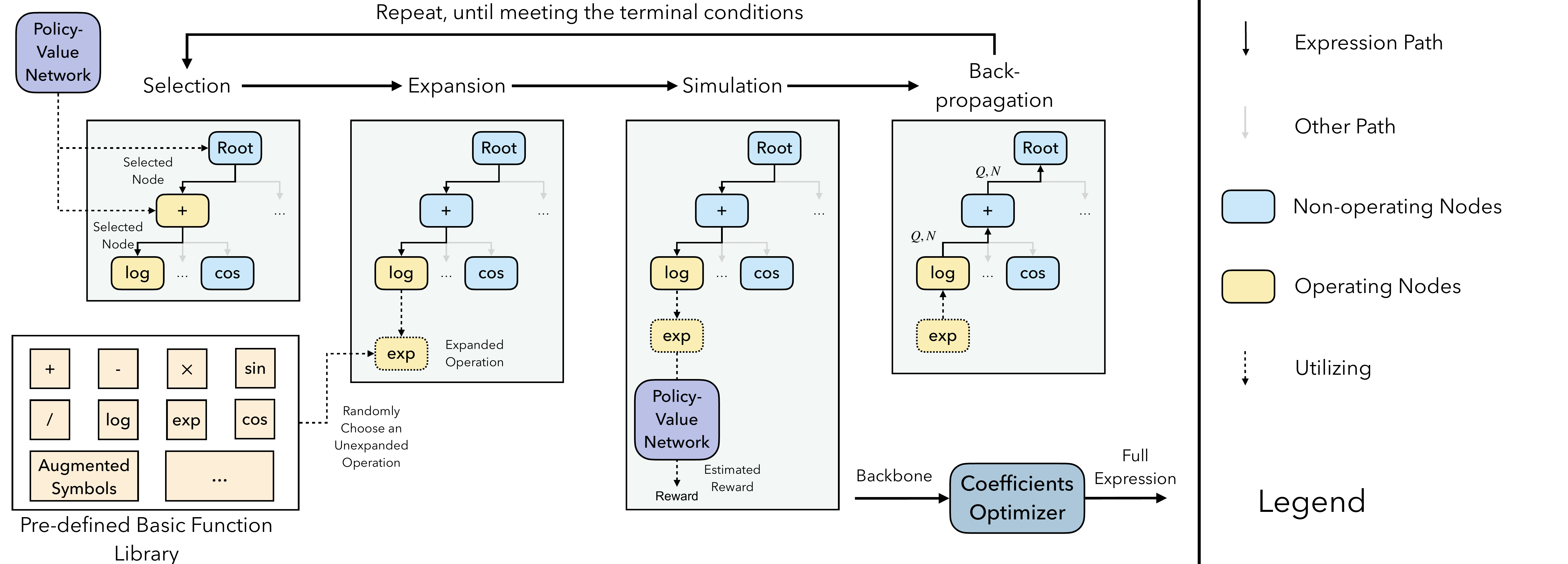}
  \caption{The overview of NEMoTS.
  }\label{NEMoTS}
\end{figure*} 

\subsection{Model Overview}
The overview of our proposed NEMoTS 
(\textbf{N}eural-\textbf{E}nhanced \textbf{Mo}nte-Carlo \textbf{T}ree \textbf{S}earch)
for time series is illustrated in Fig. \ref{NEMoTS}.

The NEMoTS comprises four main components: a pre-defined basic function library, Monte-Carlo Tree Search (MCTS), a policy-value network, and a coefficient optimizer. The first three components collaborate to form the basic structure of an expression, named "backbone", which lacks any numerical coefficients. This basic structure is then refined by the coefficient optimizer, which determines appropriate coefficients, resulting in a full expression.

Below, we first outline the collaboration among components:
\begin{itemize}
\item \textbf{Monte-Carlo Tree Search (MCTS)}: A key component of NEMoTS, MCTS is a four-phase process: selection, expansion, simulation, and back-propagation, with a function library and policy-value network playing vital roles. It creates an expression's structural "backbone," determining layout and basic functions, but not the numerical coefficients, which are set later by a coefficient optimizer.

\item \textbf{Function Library}: Crucial in MCTS's expansion phase, this library provides mathematical operations for building new nodes in the expression tree, such as addition, subtraction, and trigonometric functions.

\item \textbf{Policy-Value Network}:  Integral in MCTS's selection and simulation phases, this neural network evaluates the current expression and target time series. It selects promising nodes, assigns probabilities to operations, and scores the current state, enhancing decision-making in the search.

\item \textbf{Coefficient Optimizer}: An independent part of NEMoTS, the optimizer zeroes in on finding optimal coefficients for the MCTS-created structure using efficient numerical methods. This finalizes the analytical expression, moving from a structural to a functional form.
\end{itemize}

\subsection{Main Pipeline}
In this section, we will discuss how to generate the "backbone" of an expression 
using NEMoTS. It is important to note that each node in the tree maintains two 
variables: the total reward $Q$ and the visited count $N$.

\subsubsection{Selection}
Initially, a node designated as "Root" serves as a starting point for subsequent operations but does not contribute to expression generation.

The selection phase commences at the "Root" node and involves iterative selection of child nodes (potential mathematical operations from the pre-defined function library) until a partially expanded or unexpanded node is encountered. This process is a tree traversal employing a specific recursive strategy known as the Polynomial Upper Confidence Tree (PUCT). Formally, under tree state $S$, the PUCT score for a child node $a$ is given by:
\begin{equation}
Score(S,a)=Q(S, a) + c \cdot P(S, a) \cdot \frac{\sqrt{\sum_{b} N(S, b)}}{1 + N(S, a)}, \label{PUCT}
\end{equation}
\begin{itemize}
\item $Q(S, a)$: The average or expected reward of choosing action $a$ in state $S$, based on its performance in similar situations.
\item $P(S, a)$: The prior probability of picking action $a$ in state $S$, as estimated by the policy-value network. It gauges the potential worth of the action.
\item $N(S, a)$: The number of times action $a$ has been chosen in state $S$, measuring how much the action has been explored.
\item $c$: A constant that balances exploration (trying less-visited operations) and exploitation (using known high-rewarded operations). Higher $c$ favors exploration; the lower favors exploitation.
\item $\sum_{b} N(S, b)$ The total number of visits to all actions $b$ in state $S$, used for normalizing exploration rewards.
\end{itemize}
Utilizing Eqn. \ref{PUCT}, the child node with the highest score is chosen for further exploration until an unexplored node is reached. The path formed based on the PUCT is then converted into an expression.

\subsubsection{Expansion}
The expansion phase begins after the selection phase, focusing on a node that is either partially expanded or unexpanded, referred to as the "target node." This node is critical for adding new elements to the expression tree, central to the entire expansion process.

In this phase, a function from the library is randomly selected for expansion, following a uniform distribution. The selected operation, denoted as
$a$ , is then integrated into the tree as a new node. This node’s visit count
$N(S,a)$ and total reward 
$Q(S,a)$ are initially set to zero. This method ensures equal opportunity for each function to be chosen, promoting fairness and variety in the exploration.

After integrating the new node, the process moves to the simulation phase. This stage is vital for the overall search strategy, involving simulations to foresee possible actions and outcomes. These simulations are key to shaping future decisions. The effectiveness of the simulation phase greatly affects the selection of future nodes and the development of the expression tree.

\subsubsection{Simulation}
In MCTS, particularly within our NEMoTS framework, the simulation phase is key for assessing the potential rewards of newly expanded nodes. This phase typically follows the expansion phase and starts from the most recently added node in the expression tree. It involves a rapid simulation method, often random, and continues \textbf{until the expression path surpasses a pre-defined length, the terminal condition}. The focus here is on quick evaluation rather than deep exploration.

NEMoTS diverges from traditional random simulations, which are time-intensive. Instead, we utilize the policy-value network’s reward estimator for immediate reward estimations. This approach effectively evaluates the potential rewards of the current state and enhances the efficiency of the simulation process.

Crucially, during training, numerous random simulations are essential to provide supervised signals to the reward estimator. This ensures the policy-value network's scores are accurate and reflect real-world outcomes. This accuracy is vital for the effectiveness and precision of NEMoTS's simulation phase. The simulation concludes when the expression path reaches a predetermined length. Following the simulation phase, the generated expression path is transformed into an expression and further refined by the coefficient optimizer. The optimized expression is then assessed with the input signal as per Eqn. \ref{reward}, leading to the back-propagation phase.

\subsubsection{Back-Propagation}
The back-propagation is a critical component in the MCTS, particularly in updating the decision-making mechanism. It follows the simulation phase and initiates at the node where simulation began, often the newly expanded node, and proceeds back to the root node, referred to as the "Root."

In this phase, for each node along the path from the start node of the simulation to the root, we update both the visit count $N(S,a)$ and the total reward $Q(S,a)$. These updates are influenced by the simulation outcomes and serve to adjust $Q(S,a)$, reflecting the new average or expected reward for an action $a$ in state $S$. Concurrently, $N(S,a)$ is incremented, indicating an additional visit to that child node. The reward data obtained at the end of the simulation is vital, as it helps in evaluating the long-term strategic benefits of the node.

Back-propagation is integral to refining the overall decision-making process. It enables the algorithm to better understand and adapt to the decision space through continuous learning. This phase ensures more efficient navigation of the expression tree and improves decision-making by reinforcing successful paths and reassessing less effective ones.

The process iterates through these four phases until the expression path reaches a pre-determined threshold. At this point, a preliminary "backbone" expression is formed, which still lacks specific numerical coefficients. We then apply the Powell optimization method to this expression. This gradient-free algorithm is particularly suited for complex or non-differentiable functions, efficiently finding the function's minimum by updating search directions and conducting one-dimensional searches. This makes it an effective approach for problems where traditional gradient methods are not applicable, as discussed in \cite{Symb_Powell}.

\subsection{Model Training}
The NEMoTS training process involves two primary components: refining the policy-value network and augmenting the function library through symbolic augmentation. The objective of optimizing the policy-value network is to improve its accuracy and efficiency in aiding the MCTS process. On the other hand, expanding the function library focuses on enriching the initial set of basic functions with more complex ones. This expansion serves a dual purpose: it provides encapsulation for expressions and customizes the library to more effectively match the specific dataset.

\begin{figure*}
  \centering
  \includegraphics[width=\linewidth]{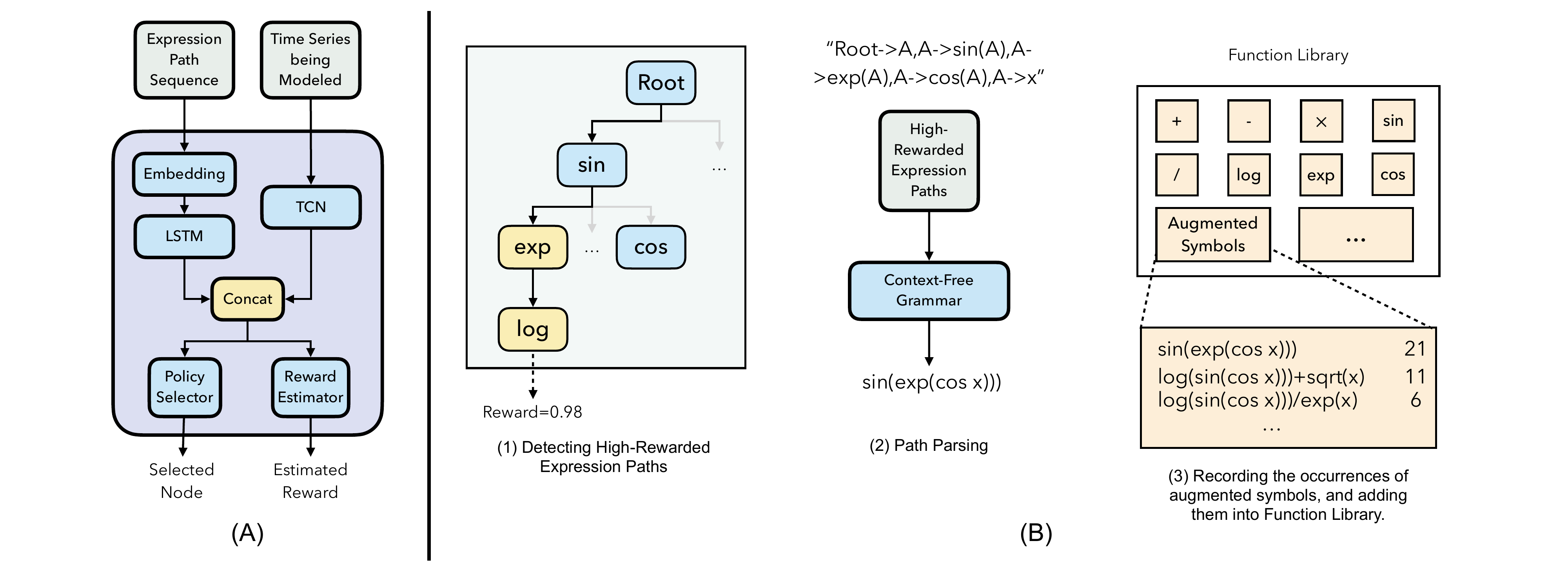}
  \caption{(A)The structure of policy-value network; and (B) The illustration of the symbolic augmentation strategy.
  }\label{NN}
\end{figure*} 

\subsubsection{Policy-Value Network}
In NEMoTS, the policy-value network serves as a black-box model, handling two types of input: the expression path sequence and the input signals, namely the time series being modeled. Its outputs are twofold: the chosen operation and the estimated reward value, both based on the expression path sequence and input signals.

For our implementation, we employ Long Short-Term Memory (LSTM) networks \cite{LSTM} to encode the expression path sequence. To process the input signal sequence, we use Temporal Convolutional Networks (TCN) \cite{TCN1}. We then concatenate the encoded representations from these two sources and use a Multilayer Perceptron (MLP) for further processing. This results in outputs for the three branches, as depicted in Fig. \ref{NN} (A).

A key aspect of our approach is optimizing the neural network. This optimization primarily focuses on designing an effective loss function that aligns with the MCTS process requirements.

\begin{itemize}
  \item \textbf{Policy Selector}. The primary goal of the policy selector in NEMoTS is to generate accurate prior probabilities \( P(S,a) \), crucial for forming a distinct probability distribution during the \( Score(S,a) \) calculation, as indicated in Eqn. \ref{PUCT}. This clarity in distribution is essential, guiding the model to selectively prioritize nodes for expansion more confidently. Accordingly, the optimization objective of this component is to minimize the Kullback-Leibler (KL) divergence between the prior probability distribution \( P(S) \) and the posterior distribution \( Score(S) \). Minimizing this divergence ensures that the model's predicted probability distribution closely mirrors the actual distribution of rewards, thereby significantly enhancing decision-making accuracy within the MCTS process.

  \begin{equation}
  Loss_{(PS)}=\sum_{a\in{A}}P(S,a)\mbox{log}\left(\frac{P(S,a)}{Score(S,a)}\right),
  \end{equation}
  where $A$ represents all valid nodes in the selection iteration.

  \item \textbf{Reward Estimator}. The reward estimator's function is to circumvent the intricate simulation phase, directly assessing the current state to produce a reward value. This makes it essentially a regression model. To train this component of the neural network effectively, we focus on minimizing the Mean Squared Error (MSE) between the output $\mathcal{R}'$ of the reward estimator and the simulated reward value $\mathcal{R}$. This minimization ensures that the reward estimator's predictions are as close as possible to the actual reward outcomes, thereby refining the model’s efficiency in estimating rewards without the need for complex simulations:
  \begin{equation}
  Loss_{(RE)}=(\mathcal{R}'-\mathcal{R})^{2}
  \end{equation}
  \end{itemize}
  
  Thus, optimizing the policy-value network equates to minimizing the 
  loss of the above two parts:
  \begin{equation}
  Loss=\theta_{1}Loss_{(PS)}+\theta_{2}Loss_{(RE)}.
  \end{equation}
  $\theta_{1},\theta_{2}$ are coefficients used to balance these terms.

\subsubsection{Symbolic Augmentation Strategy}

In our study, we observed that relying solely on elementary functions in the function library often proved inadequate for accurately representing complex nonlinear dynamical systems in symbolic regression tasks. Consequently, the incorporation of more sophisticated composite functions became essential for precise time series representation. Contrary to previous approaches that depended on random selection in the MCTS simulation phase to identify high-reward expressions from a plethora of randomly generated composite functions \cite{Symb_SPL}, our method takes a different path. We substituted the traditional random simulation with neural networks, which initially lacked the specialized ability to create specific composite functions for certain samples.

During the training phase, we implemented a strategy similar to frequent pattern mining \cite{Symb_FreqItem}. This involved tracking expression paths and their corresponding high-reward composite functions. Typically, these expressions with high rewards emerged from random combination selections. According to the law of large numbers \cite{Symb_lln}, among these paths, some consistently received high rewards. After training, we analyzed the frequency of these high-reward paths. The most recurrent paths were deemed as optimal composite functions for our dataset and subsequently incorporated into our function library. This addition significantly enhanced our model’s ability to represent time series and boosted the accuracy of modeling complex systems.

For practical applications, a novel element, termed 'augmented symbols', is introduced into the function library. During the expansion phase in MCTS, if the model selects this 'augmented symbol', it triggers a secondary sampling process. This process is based on a probability distribution reflecting the occurrence frequency of the top $k$ high-reward composite functions. Under this mechanism, a higher frequency of a specific 'augmented symbol' in the dataset indicates its more prevalent usage during training, suggesting a better fit for the current dataset, as illustrated in Fig. \ref{NN} (B). This approach markedly improves the model's adaptability to specific dataset characteristics, thus enhancing both the accuracy and efficiency of the symbolic regression process.

\section{Experiments}
We conducted experiments on three real-world datasets to answer the following four questions:
\begin{itemize}
\item Q1: How is the fitting ability and efficiency of NEMoTS?
\item Q2: Are the expressions generated by NEMoTS reliable?
\item Q3: What enhancements in performance and efficiency do NEMoTS achieve through improvements?
\item Q4: How can the interpretability of NEMoTS be analyzed?
\end{itemize}

To address these questions, we will conduct four sets of experiments: fitting 
ability performance, ablation studies, extrapolation analysis, and case studies.

\subsection{Datasets and Pre-Processing}
In our research, we selected three univariate datasets for analysis:
\begin{itemize}
\item Weighted Influenza-Like Illness Percentage (WILI);
\item Australian Daily Currency Exchange Rates (ACER);
\item Atmospheric Pressure (AP).
\end{itemize}
The datasets used in this study include ILI, Exchange rate, and Weather, as provided by previous research \cite{Informer,Autoformer}. We extracted relevant columns from these publicly available datasets. For symbolic regression tasks, we modified the ACER and AP datasets to include only the first 1000 timestamps, creating sub-datasets due to their time-intensive nature. The entire WILI dataset was utilized for comprehensive analysis owing to its smaller size.

\subsection{Evaluation Metrics}
We evaluate algorithm performance using the coefficient of determination (\(R^2\)) and the correlation coefficient (CORR), which counteract the influence of data size. Efficiency is assessed through the Average Time Cost per sample in seconds (ATC), reflecting the algorithms' fitting ability and computational speed, useful for algorithm selection and optimization.

Given the significant random search in symbolic regression algorithms, we use actual time cost for efficiency evaluation. Though CPU tests were isolated to minimize external process interference, some variability in time cost due to computing environment fluctuations is expected. Despite their limitations, these measurements offer a general insight into the models' efficiency.
For both \(R^2\) and CORR, values nearing 1 indicate smaller regression error and a trend closer to actual values, respectively.

\subsection{Fitting Ability Performance} \label{sec:overall}
To address Q1, we focused on assessing the overall fitting ability. We used a sliding sampling method to divide time series data into samples of 36 and 72 time steps, for shorter and longer series respectively. In NEMoTS neural network training, 10\% of each dataset was used for learning parameters, with the remaining 90\% for testing. Importantly, NEMoTS’s network is trained on the states processed by MCTS rather than directly on time series patterns, which obviates the need for a validation set. Table \ref{tab:main_res} provides a comparative performance summary. NEMoTS, once fully trained, excelled in 16 out of 18 metrics across six experimental sets involving three datasets and ranked second in the other two, highlighting its overall superiority. Our subsequent analysis will concentrate on two main aspects: performance, evaluated using the coefficient of determination \(R^2\) and correlation coefficient CORR, and efficiency, assessed by average time cost per sample.

\subsubsection{Baselines}
We will compare with the following methods:
\begin{itemize}
\item Genetic Programming (GP) evolves through selection, crossover, and mutation to find optimal data-fitting combinations, implemented via gplearn's interface \cite{genetic_programming, gplearn2024}.
\item Multiple Regression Genetic Programming (MRGP) enhances GP by decoupling and linearly combining program sub-expressions through multiple regression, outperforming traditional GP and multiple regression \cite{Symb_MRGP}.
\item Bayesian Symbolic Regression (BSR) approaches symbolic regression by sampling symbolic trees from a posterior distribution using Markov Chain Monte Carlo \cite{Symb_BSR}.
\item Physics Symbolic Optimization (PhySO) recovers analytic expressions from physical data using deep reinforcement learning and maintains physical unit consistency by learning unit constraints \cite{Symb_PhySO}.
\item Symbolic Physics Learner (SPL) employs a naive Monte-Carlo Tree Search method for symbolic regression \cite{Symb_SPL}.
\end{itemize}
For the coefficient-less backbones from these methods, we fit using least squares. Given the randomness in symbolic regression algorithms, which affects performance on certain samples, we exclude obviously abnormal metrics for fair comparison.

\begin{table}[htbp]
  \setlength\tabcolsep{3pt}
  \centering
  \scriptsize
  \caption{Fitting Results. Both $R^{2}$ and CORR are dimensionless metrics, and ATC (Average Time Cost) by seconds.} 
  \label{tab:combined_performance}
  \begin{tabular}{c|c|c|ccccccc}
    \toprule
    \multicolumn{3}{c|}{Baselines} & GP & MRGP & BSR & PhySO & SPL & NEMoTS \\
    \cmidrule{1-9}
    Dataset& Seq. Length & Metrics & \multicolumn{6}{c}{} \\
    \midrule
    \multirow{6}{*}{WILI} 
    &\multirow{3}{*}{36} & $R^{2}$ & 0.302 & 0.377 & 0.541 & 0.488  & \textbf{0.937} & 0.923 \\
    & & CORR & 0.593 & 0.621 & 0.603 & 0.640 & \textbf{0.951} & 0.940 \\
    & & ATC    & 93.35 & 203.44 & 113.23 & 103.51 & 223.25 & \textbf{28.13} \\
    \cmidrule{2-9}
    &\multirow{3}{*}{72} & $R^{2}$ & 0.287 & 0.267 & 0.089 & 0.375  & 0.863 & \textbf{0.890} \\
    & & CORR & 0.502 & 0.513 & 0.287 & 0.447 & 0.912 & \textbf{0.930} \\
    & & ATC    & 90.51 & 198.51 & 57.72 & 106.53 & 231.01 & \textbf{42.18} \\
     \cmidrule{1-9}
     \multirow{6}{*}{ACER} 
&\multirow{3}{*}{36} & $R^{2}$ & 0.133 & 0.318 & 0.327 & 0.616 & 0.752 & \textbf{0.842} \\
& & CORR & 0.215 & 0.422 & 0.541 & 0.701 & 0.838 & \textbf{0.857} \\
& & ATC    & 68.49 & 133.52 & 110.29 & 99.63 & 269.76 & \textbf{30.34} \\
\cmidrule{2-9}
&\multirow{3}{*}{72} & $R^{2}$ & 0.081 & 0.497 & 0.238 & 0.663 & 0.609 & \textbf{0.738} \\
& & CORR & 0.178 & 0.531 & 0.364 & 0.715 & 0.780 & \textbf{0.831} \\
& & ATC   & 66.27 & 161.52 & 66.70 & 97.85 & 296.41 & \textbf{44.23} \\
    \cmidrule{1-9}
     \multirow{6}{*}{AP} 
&\multirow{3}{*}{36} & $R^{2}$ & 0.769 & 0.780 & 0.657 & 0.231 & 0.825 & \textbf{0.931} \\
& & CORR & 0.716 & 0.707 & 0.628 & 0.615 & 0.869 & \textbf{0.955} \\
& & ATC    & 123.83 & 192.45 & 133.51 & 123.51 & 202.92 & \textbf{31.41} \\
\cmidrule{2-9}
&\multirow{3}{*}{72} & $R^{2}$& 0.171 & 0.397 & 0.358 & 0.173 & 0.858 & \textbf{0.916} \\
& & CORR & 0.378 & 0.566 & 0.461 & 0.286 & 0.906 & \textbf{0.943} \\
& & ATC    & 143.65 & 203.45 & 75.13 & 125.47 & 217.43 & \textbf{39.54} \\

    \bottomrule
  \end{tabular}\label{tab:main_res}
\end{table}

\subsubsection{Performance}
The coefficient of determination $R^{2}$, a ratio of model error to average error, is a dimensionless metric assessing model fitting ability. Quantitatively, SPL and NEMoTS significantly surpass other models in the $R^{2}$ metric. SPL shows an average 203.04\% improvement over models like GP, MRGP, BSR, and PhySO in $R^{2}$, while NEMoTS averages 229.21\% improvement over all but SPL. This highlights SPL and NEMoTS's superior model fitting capabilities.
The correlation coefficient (CORR) assesses the consistency between fitted and actual values. In this metric, SPL and NEMoTS also demonstrate considerable advantages. SPL achieves a 96.22\% average improvement in CORR over the aforementioned models, and NEMoTS shows a 103.65\% improvement over all but SPL. These results underscore the effectiveness of SPL and NEMoTS in aligning model predictions with actual data trends.

SPL and NEMoTS's fitting proficiency partly stems from incorporating the MCTS algorithm. MCTS's design calculates historical returns to efficiently select high-rewarded operations and expressions. It also abstracts expressions as tree structures, ensuring validity and reducing numerical problems.
NEMoTS outperforms SPL, owing to its policy-value network that learns from extensive data, enabling more focused selection and expansion phases in search of quality expressions. However, MCTS's inherent randomness can impact model performance stability. Despite this, the MCTS-based design of NEMoTS and SPL offers robust fitting ability for symbolic regression models.

\subsubsection{Efficiency}
We assessed algorithm efficiency by evaluating the average time cost per sample. The results show NEMoTS with a substantial reduction in average time cost compared to other methods (GP, MRGP, BSR, PhySO, SPL), achieving about a 68.06\% improvement. This efficiency gain primarily arises from incorporating the policy-value network. Unlike other models, NEMoTS bypasses numerous simulations and search steps, using its neural network for direct assessment of the current expression. This approach, by eliminating the need for time-intensive simulations and searches, markedly improves NEMoTS's efficiency. Consequently, NEMoTS exhibits a significant competitive edge in time efficiency, proving advantageous for large-scale or time-sensitive tasks.

\subsection{Extrapolation}

To answer Q2, we undertake an extrapolation task, commonly known as short-term prediction, which is a pivotal aspect of time series analysis. The fundamental idea is to evaluate whether an expression, derived through symbolic regression, can accurately predict the future evolution of a time series. Successfully doing so would demonstrate that the expression has adeptly captured the core pattern inherent in the time series. This critical evaluation allows us to substantiate the interpretative reliability of our model, offering insights into its ability to decipher and project data trends.

Practically, we will implement this extrapolation task on the same three datasets previously mentioned. In each case, we will analyze time series data covering 30 time steps, with the objective of forecasting the subsequent 6 time steps. This methodology is designed to rigorously test the model's proficiency in short-term forecasting. By doing so, we aim to comprehensively evaluate the model’s capacity to not only understand but also accurately project the underlying patterns of time series data. This approach is particularly useful in determining the model’s effectiveness in navigating and interpreting complex data sequences, thereby providing a robust assessment of its predictive capabilities and reliability in real-world scenarios.

In this section, we will use the following time series prediction methods to compare:
\begin{itemize}
    \item Auto-Regressive Integrated Moving Average (ARIMA) \cite{RW_arima};
    \item Support Vector Regression (SVR) \cite{SVR};
    \item Recurrent Neural Network (RNN) \cite{RNN1};
    \item Temporal Convolutional Networks (TCN) \cite{TCN1};
    \item Neural Basis Expansion Analysis (NBeats) \cite{NEATS}. 
\end{itemize}
It is important to note that we do not intend to introduce overly complex prediction models in this context. This is because the purpose of the extrapolation task is merely to verify whether the derived expression captures the intrinsic dynamics of the time series. The primary objective of this study is to offer insights into interpretability, rather than to focus on prediction. 

\begin{table}[htbp]
  \setlength\tabcolsep{3pt}
  \centering
  \scriptsize
  \caption{Extrapolation Performance} 
  \label{tab:combined_performance}
  \begin{tabular}{c|c|ccccccc}
    \toprule
    \multicolumn{2}{c|}{baselines} & ARIMA & SVR & RNN & TCN& NBeats & NEMoTS \\
    \cmidrule{1-8}
    Dataset & Metrics & \multicolumn{6}{c}{} \\
    \midrule
    \multirow{2}{*}{WILI}
    & $R^{2}$ &-1.512  & -1.085 & 0.231 & 0.214 & 0.453& \textbf{0.617} \\
    & CORR  &0.234  & 0.181 & 0.313 & 0.307 & 0.312& \textbf{0.566} \\
    \cmidrule{1-8}
    \multirow{2}{*}{ACER}
    & $R^{2}$ &-1.187  & -0.943 & 0.435 & 0.115 & 0.393& \textbf{0.479} \\
    & CORR  &0.126  & 0.228 & 0.133 & 0.203 & 0.465& \textbf{0.617} \\
    \cmidrule{1-8}
    \multirow{2}{*}{AP}
    & $R^{2}$ &-1.433  & -1.854 & 0.215 & 0.336 & 0.461& \textbf{0.628} \\
    & CORR  &0.445  & 0.423 & 0.417 & 0.518 &\textbf{0.658}& 0.629 \\
    \bottomrule
    
  \end{tabular}
\end{table}

Overall, the NEMoTS model demonstrates superior performance in most scenarios, particularly excelling in the $R^2$ metric, where it achieves the highest scores across all three datasets. In contrast, the ARIMA and SVR models generally exhibit poor performance, especially in the $R^2$ metric, where these models show negative values in most datasets, indicating lower prediction accuracy. On the AP dataset, all models exhibit relatively high CORR values, suggesting that their predictions are more closely correlated with actual outcomes.

The performance of all models on two key metrics is only moderately satisfactory, mainly due to two factors: the challenging nature of the datasets, which are non-stationary and non-periodic, making prediction difficult; and the limited training data, with only 10\% of samples used from 1000 timestamps, leading to insufficient training of neural network-based methods. These issues combine to limit the models' ability to accurately forecast future trends.

Despite these challenges, NEMoTS showcases superior performance when compared to other prediction models. This is indicative not just of its proficiency in prediction and extrapolation tasks but, more crucially, of its ability to grasp the essential evolutionary traits of time series data. The capacity of NEMoTS to accurately identify these underlying dynamics, despite the time series being subject to a multitude of influencing factors, speaks volumes about its effectiveness. This success in capturing the intrinsic patterns of time series data is a testament to the model's robustness in deciphering complex data structures and the reliability of the insights it generates, thereby affirming its utility in practical applications where understanding and predicting data trends are vital.

\subsection{Ablation Studies}
\begin{figure}
  \centering
  \includegraphics[width=\linewidth]{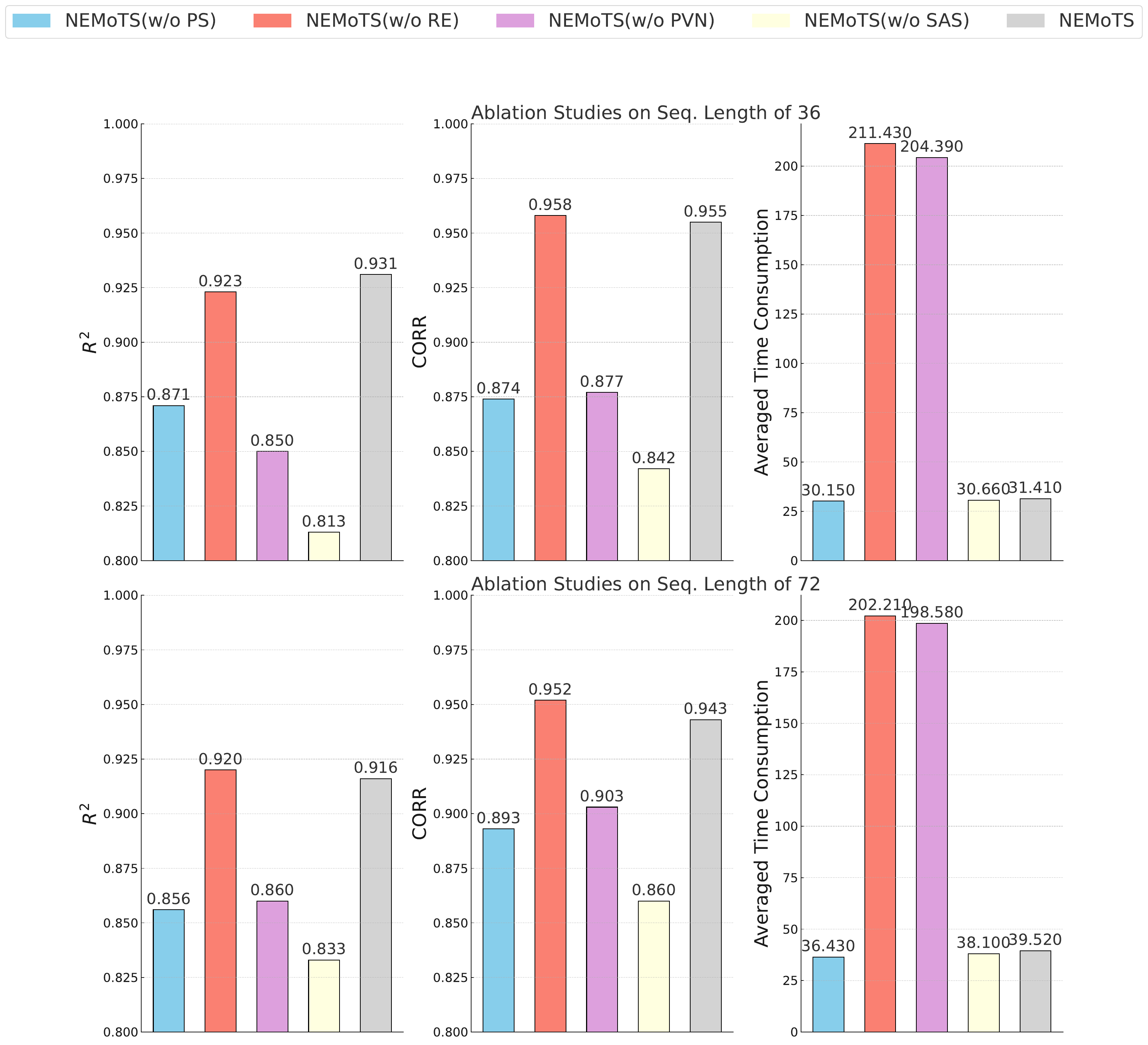}
  \caption{The results of ablation studies.
  }\label{ablation}
\end{figure}

To address Q3, we carried out ablation studies by individually omitting the policy-value network and the symbolic augmentation strategy from NEMoTS to evaluate their impacts on efficiency and performance. Specifically, we tested the model under four scenarios: 1) NEMoTS without the policy selector (NEMoTS(w/o PS)), utilizing the UCB score for node selection; 2) NEMoTS without the reward estimator (NEMoTS(w/o RE)), employing a random strategy for reward simulation; 3) NEMoTS lacking the entire policy-value network (NEMoTS(w/o PVN)); and 4) NEMoTS minus the symbolic augmentation strategy (NEMoTS(w/o SAS). The experiments were conducted on the AP dataset, with sequence lengths of 36 and 72. Results of the ablation study are displayed in Fig. \ref{ablation}.

\subsubsection{Policy Selector}
The policy selector in NEMoTS plays a critical role in evaluating the expression sequence and time series data from the root node, essential for guiding the selection of child nodes in Monte-Carlo Tree Search (MCTS). It effectively narrows the search space, enhancing efficiency and precision.
Ablation study results emphasize its significance. Its removal results in a 6.45\% decrease in the coefficient of determination (\(R^2\)) and a 6.89\% reduction in the correlation coefficient (CORR). This suggests that without the policy selector, MCTS is less efficient at identifying optimal operations, negatively affecting overall performance.
Notably, the absence of the policy selector also leads to a lower average time cost per sample. In the ablation model without the policy selector, the standard prior distribution is replaced by a uniform distribution, thereby reducing the need for neural network operations and shortening processing time.

\subsubsection{Reward Estimator}
The reward estimator is pivotal for evaluating the current state's reward score, thereby bypassing the need for extensive simulations typical in traditional MCTS. Our ablation study shows that while its removal slightly improves model performance, it substantially increases the time cost.
This higher time cost is due to traditional MCTS's dependence on numerous, time-intensive simulations. The slight improvement in performance may be because neural network-based reward estimations aren't as precise as those from simulations, which could impact the model's precision. Essentially, the neural network offers a faster but potentially less accurate scoring method compared to traditional simulations.
Therefore, removing the reward estimator leads to a minor improvement in performance but a significant increase in time cost. This trade-off underlines the importance of balancing performance gains and time efficiency in practical scenarios, especially when deciding on the use of a reward estimator.

\subsubsection{Policy-Value Network}
In the Monte-Carlo Tree Search (MCTS) process, each element of the policy-value network – the policy selector and the reward estimator – is vital, greatly influencing decision accuracy, expansion efficiency, and precision in reward assessment. Without the policy selector, there's a decline in efficiency for selecting optimal operations. Conversely, removing the reward estimator could enhance performance but at the expense of increased time cost.
The removal of the entire policy-value network underscores the combined impact of these components. This exclusion can result in varied model performance, highlighting the need for a balanced interplay among the different elements in the model.

\subsubsection{Symbolic Augmentation Strategy}
Symbolic augmentation strategy enhances the function library by incorporating high-reward composite functions developed during training. These functions capture specific dataset patterns more comprehensively than basic elementary functions, offering a fuller understanding of time series characteristics.In standard Monte-Carlo Tree Search (MCTS), identifying complex patterns is often difficult. However, the symbolic augmentation strategy aids in recognizing and using these complex patterns more effectively, significantly boosting model performance. Without this strategy, there's a potential for underutilization of these intricate patterns, which could lead to a marked decrease in performance. In summary, the symbolic augmentation strategy is key in NEMoTS, as it includes high-reward composite functions in the function library. This enhances the model's ability to identify and articulate complex time series patterns, thus improving accuracy and efficiency in both analysis and prediction.

\subsection{Case Study}

\begin{figure}
  \centering
  \includegraphics[width=\linewidth]{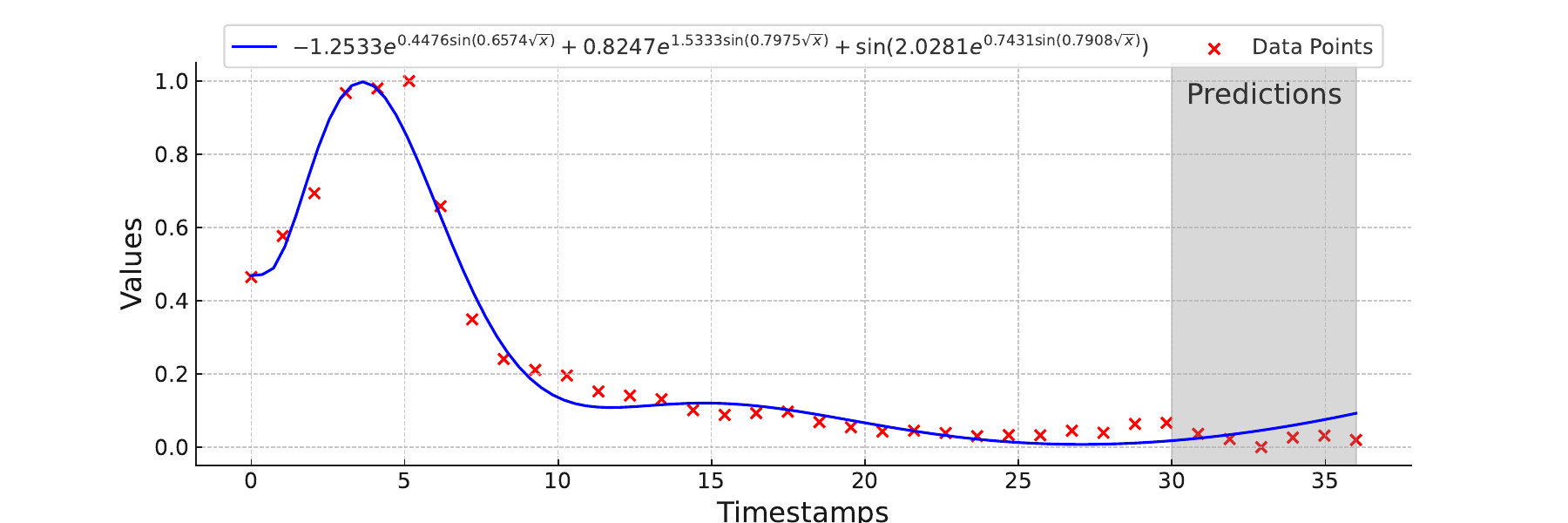}
  \caption{Case study, where the white background represents the data used for fitting, and the gray part represents the data not used during fitting, which is the prediction part.
  }\label{casesty}
\end{figure} 
To address Q4, we compare NEMoTS-derived symbolic regression expressions with actual dataset data, as shown in Fig. \ref{casesty}, with additional case studies in Appendix \ref{sec:casestudies} and Fig. \ref{fig:casestudies}. These visualizations demonstrate NEMoTS's effectiveness in fitting complex real data with succinct mathematical models, capturing time series trends. This confirms NEMoTS's capability in fitting intricate data and its efficiency in trend extraction and representation, providing insightful analysis of time series.
Additionally, our analysis includes predictive assessments on future data not used in the fitting, shown in sections with a gray background in the figures. NEMoTS not only accurately fits existing data but also forecasts future trends reliably. This indicates NEMoTS's strong predictive power, enhancing the value of its identified expressions.
The ability of NEMoTS to accurately predict future trends signifies its interpretability and reliability. This accuracy is crucial for deeper understanding and forecasting of time series data, especially in domains that demand precise data prediction and interpretation.

\section{Conclusion}
In this study, we apply symbolic regression techniques to time series analysis, improving its interpretability by extracting analytical expressions from time series data. To tackle the large search space in symbolic regression, we adopt Monte-Carlo Tree Search (MCTS), which narrows down the search space and ensures expression validity through structured constraints. We enhance the method's efficiency and generalization by integrating neural networks, which guide MCTS and replace conventional simulations, boosting efficiency and learning capabilities. Additionally, our Symbolic Augmentation Strategy captures and utilizes common composite functions, enhancing the model's fitting ability. Our extensive tests on three real-world datasets demonstrate our method's superior performance, efficiency, reliability, and interpretability in time series analysis.

% Symbolic regression techniques, in particular, are adept at 
% approximating evolution patterns more accurately due to their lack of a 
% priori constraints \cite{}.

% These frameworks include Auto-Regressive 
% Integrated Moving Average (ARIMA),
% Kalman Filter \cite{Kalman}, Gradient Boosting Decision Tree (GBDT) \cite{GBDT1,GBDT2}, 
% representives of 
% statistical learning methods, and deep leanring 
% methods like Recurrent Neural Networks (RNN) \cite{RNN1,RNN2,RNN3}, 
% Temporal Convolutional Networks (TCN) \cite{TCN1,TCN2}, and various 
% self-attention based methods (Transformers) \cite{Transformer,Informer,Autoformer,TransformerTSSurvey}.
% They mainly focus on \textbf{how} will the time series evolves, but ignore \textbf{why} 
% the time series will evolve like this and evolution patterns it has.

% \begin{acks}
% \end{acks}

\bibliographystyle{ACM-Reference-Format}
\bibliography{sample-base}

\appendix

\section{Related Works}
\subsection{Symbolic Regression}
Symbolic Regression (SR) emerges as a sophisticated technique in regression analysis, 
uniquely characterized by its ability to simultaneously identify both the parameters 
and the functional forms of equations that best describe given datasets. This method 
stands apart from traditional regression approaches, such as linear or quadratic 
regression, by offering a more holistic and adaptable framework for data 
analysis \cite{Symb_intro1,SRRW1,SRRW2}.

Central to SR's methodology is its data-driven nature, which operates independently 
of preconceived models or theories about the system under investigation \cite{SRRW3}. 
This characteristic is particularly advantageous when dealing with datasets that encompass 
ambiguous or complex relationships. SR's capacity to unearth these intricate associations 
not only provides innovative solutions to challenging problems but also fosters a deeper 
comprehension of systems that are only partially understood. Furthermore, SR's prowess 
in generating closed-form mathematical expressions renders it an invaluable asset in the 
realm of generalizable AI. Its compatibility with various modeling tools, such as finite 
element solvers, underscores its versatility and broad applicability \cite{SRRW4}.

One of the most notable achievements of SR is its ability to rediscover and validate 
fundamental physical laws, exemplified by its replication of Newton's law of gravitation 
through purely data-driven means \cite{SRRW5,SRRW6}. This capacity underscores 
SR's potential in empirically grounding theoretical constructs. However, it's important 
to acknowledge the challenges inherent in this method. The risk of deriving spurious 
results due to oversimplified datasets or the absence of robust evaluation metrics is a 
notable concern \cite{SRRW7}. SR shows particular efficacy in analyzing complex, nonlinear 
dynamic systems, distilling governing expressions directly from observational data. Despite
these strengths, SR's effectiveness can be compromised by factors such as data scarcity, 
low fidelity, and noise. Nevertheless, the application of Bayesian methodologies has shown 
promise in mitigating these limitations \cite{SRRW9}. Compared to several machine learning 
models, SR has demonstrated superior performance, especially in scenarios involving small 
datasets. However, a significant drawback of SR is its computational intensity, as the 
evaluation of numerous potential equations can be time-consuming. This aspect makes SR 
more suitable for scenarios with a limited number of input parameters \cite{SRRW8}.

\subsection{Monte-Carlo Tree Search}
Monte-Carlo Tree Search (MCTS) is a prominent method in artificial intelligence for 
making optimal decisions by sampling randomly in the decision space and constructing
 a search tree based on the outcomes \cite{Symb_MCTS_BG}. This approach has significantly impacted AI, 
 particularly in fields that can be modeled as trees of sequential decisions, like 
 games and planning problems \cite{Symb_AlphaGo,Symb_AlphaZero}.

A typical MCTS consists of four main stages:
\begin{enumerate}
    \item \textbf{Selection}: The process begins at the root node and involves 
    selecting successive child nodes until a leaf node is reached. This selection 
    is based on a tree policy, typically the Upper Confidence Bound (UCB) applied 
    to trees, which balances exploration and exploitation. The UCB formula takes into 
    account both the average reward of the node and the number of times it has been visited;
    \item \textbf{Expansion}: Once a leaf node is reached, one or more child nodes 
    are added to expand the tree, depending on the available actions. This step is 
    crucial for exploring new parts of the search space; 
    \item \textbf{Simulation}: This involves simulating a play from the newly added nodes 
    to the end of the game using a default or random policy. The simulation phase is 
    where MCTS diverges from traditional tree search methods, as it involves playing 
    out random scenarios to get an estimate of the potential outcome from the current state.
    \item \textbf{Back-propagation}: In the final phase, the results of the simulation 
    are propagated back up the tree. The nodes visited during the selection phase are 
    updated with the new information, typically involving updating their average reward 
    based on the outcome of the simulation and incrementing the visit count.
\end{enumerate}

MCTS is fundamentally a decision-making algorithm grounded in search and probability rather than a conventional machine learning algorithm. However, due to its complex heuristic rules, the algorithm encounters several challenges: (1) Its intricate simulation process requires numerous simulations, resulting in high complexity and low efficiency; (2) As a decision-making algorithm, MCTS operates only at the instance level and lacks the capacity to learn from large-scale data, thus missing the opportunity for inductive learning from extensive datasets to enhance performance. This paper addresses these two critical issues. We adapt MCTS to the specific needs of symbolic regression tasks, focusing on discovering expressions while also aiming to enhance MCTS's efficiency in this domain. Additionally, we enable MCTS to acquire learning capabilities, allowing it to optimize itself with vast amounts of data.

\section{Discussing Complexity: NEMoTS and Naive MCTS}

\subsection{Overview}
This part talks about how complex NEMoTS is compared to simple MCTS in tasks about symbolic regression. We think about cases where the longest expression is \(L\) and there are \(|A|\) symbols to choose from. Since both NEMoTS and simple MCTS use chance, it's hard to say exactly how complex they are for a certain problem. So, we look at the most complex it could be.

% \subsubsection{Testing Complexity}
See Section \ref{sec:overall} for tests on how complex these methods are.

\subsection{Simple MCTS Complexity}
\newpage

MCTS is a probability-based way to search, where each part of the tree is a different symbol action. If the longest expression is \(L\) and there are \(|A|\) symbols, the most nodes you could have is \(\sum_{i=0}^{L}|A|^{i}=\frac{|A|^{L+1}-1}{|A|-1}\). From this, we understand:

\begin{itemize}
    \item \textbf{Space Complexity}: Mostly about how many nodes are in the tree. If the tree grows a lot, it could have almost \(|A|^{L+1}\) nodes, making the space complexity \(\mathcal{O}(|A|^L)\).
    
    \item \textbf{Time Complexity}: We need to look at each main part of MCTS. The \textbf{selection phase} takes \(\mathcal{O}(L)\) time as it picks a path from start to end. The \textbf{expansion phase} can add a new part each time, taking \(\mathcal{O}(1)\) time. The \textbf{simulation phase} depends on how many simulation steps \(S\) there are, and we think it's \(\mathcal{O}(S)\). Lastly, the \textbf{backpropagation phase} updates the path from the current part to the start, also taking \(\mathcal{O}(L)\) time. All these parts together mean each round takes about \(\mathcal{O}(2L+S)\) time.
\end{itemize}

\subsection{NEMoTS Complexity}
NEMoTS, which adds neural networks to MCTS, has the same space complexity as MCTS, \(\mathcal{O}(|A|^L)\). The neural network's complexity is usually seen as a constant \(\mathcal{O}(1)\), not linked to MCTS's tree depth or how many choices each part has. NEMoTS is better because it speeds up the \textbf{simulation phase}, so we think its time complexity is \(\mathcal{O}(2L)\). When we compare it to simple MCTS, the big difference is that the \(\mathcal{O}(S)\) part gets smaller, so \(\mathcal{O}(2L+S)-\mathcal{O}(2L)=\mathcal{O}(S)\).

% In cases where the longest expression has at most \(L\) actions and 
% considering MCTS's rules, the number of simulation steps \(S\) is much bigger
% than \(L\). In simple MCTS, this means doing \(S\) random tests for each new 
% part. 

In our tests with \(L=20\) and \(S=200\), adding neural networks to NEMoTS can cut down the number of steps by about 83.3\% with these settings, showing why NEMoTS is more efficient.

\section{Case Studies}\label{sec:casestudies}
\begin{figure}
  \centering
  \includegraphics[width=\linewidth]{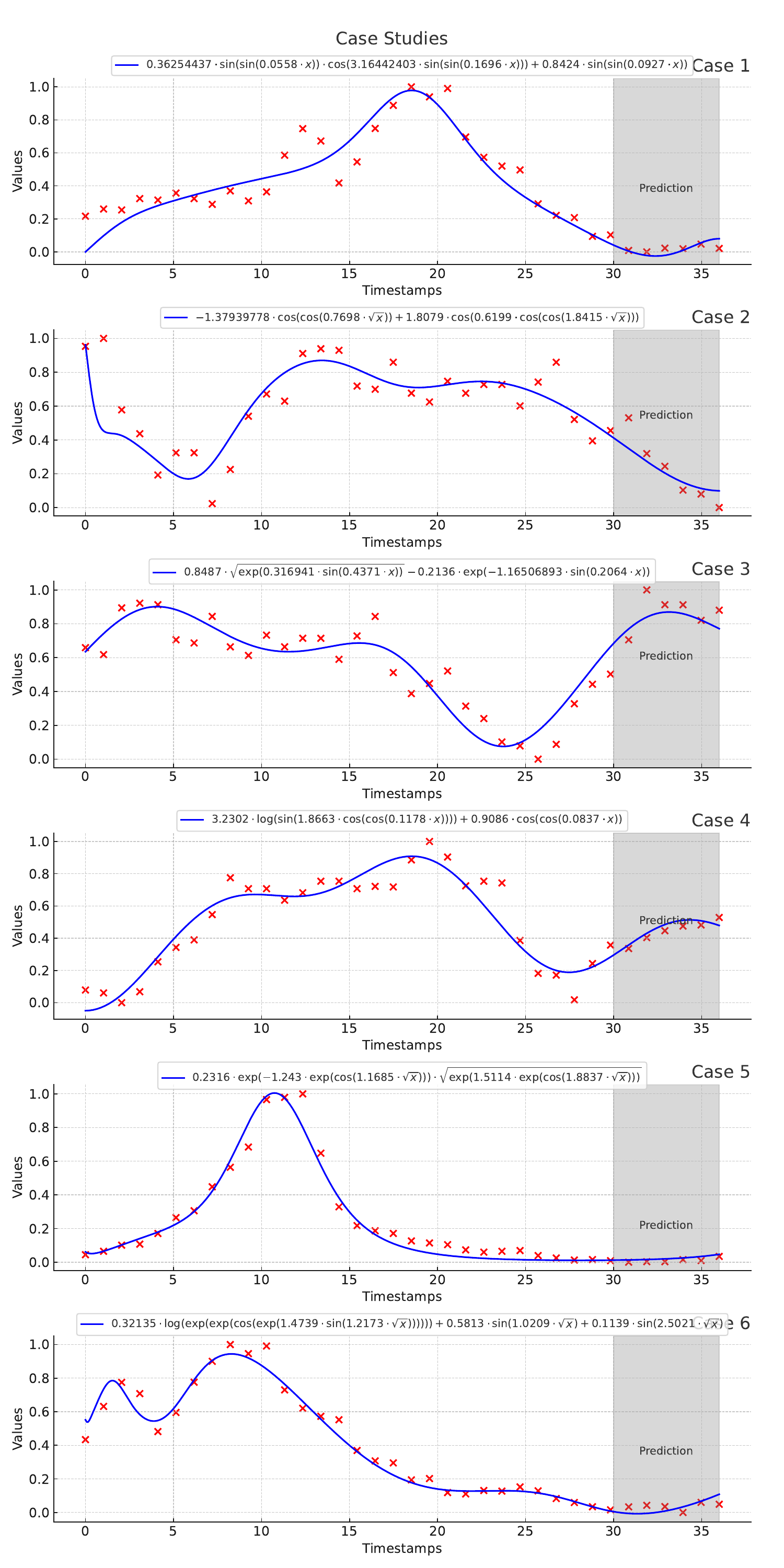}
  \caption{Case studies
  }\label{fig:casestudies}
\end{figure}

More case studies are shown in Fig. \ref{fig:casestudies}.

\section{Algorithm Process}
The detailed algorithm process is shown as Algorithm \ref{algo2}.

\begin{algorithm} 
% \label{algo2}
\DontPrintSemicolon
\caption{Process of Neural-Enhanced Monte-Carlo Tree Search (NEMoTS)}\label{algo2}

\KwInput{Time series data $\mathcal{D}$, function library, policy-value network}
\KwOutput{Full expression}

\SetKwProg{Proc}{Procedure}{:}{\KwRet{}}
\SetKwFunction{FSelect}{Select}
\SetKwFunction{FExpand}{Expand}
\SetKwFunction{FSimulate}{Simulate}
\SetKwFunction{FBackPropagate}{Back-Propagate}
\SetKwFunction{FPuct}{PUCT}
\SetKwFunction{FOptimizeCoefficients}{Optimize Coefficients}
\SetKwFunction{FSymbolAugmentationStrategy}{Symbolic Augmentation Strategy}

\Proc{NEMoTS}{
    Initialize root node as "Root"\;
    \While{not the terminal condition}{
        $leaf \gets$ \FSelect{$root$}\;
        $child \gets$ \FExpand{$leaf$}\;
        $reward \gets$ \FSimulate{$child$}\;
        \FBackPropagate{$child, reward$}\;
        \FSymbolAugmentationStrategy{$child, reward$}\;
        % $Root \gets leaf$\;
    }
    $backbone \gets$ Generate backbones from $root$\;
    $fullExpression \gets$ \FOptimizeCoefficients{$backbone$}\;
}

\SetKwProg{Fn}{Function}{:}{\KwRet{}}
\Fn{\FSelect{$node$}}{
    \While{$node$ not fully expanded}{
        $node \gets$ \FPuct{$node$}\;
    }
    \KwRet{$node$}
}

\Fn{\FExpand{$node$}}{
    % Get expansion probabilities using policy-value network
    Randomly select an operation from function library\;
    \If{select operation is the augmented symbols}{
        Perform secondary sampling from augmented symbols\;
    }
    Add new child node to $node$ with selected operation\;
    \KwRet{new child node}\;
}

\Fn{\FSimulate{$node$}}{
    Evaluate $node$ using reward estimator\;
    \KwRet{evaluation reward}\;
}

\Fn{\FBackPropagate{$node, reward$}}{
    \While{$node$ not null}{
        Update $node$'s total reward $Q$ and visit count $N$\;
        $node \gets$ parent of $node$\;
    }
}

\Fn{\FPuct{$node$}}{
    Calculate PUCT score for each child of $node$\;
    \KwRet{child with highest PUCT score}\;
}

\Fn{\FOptimizeCoefficients{$expression$}}{
    Optimize $expression$ using gradient-free methods (e.g., Powell's method)\;
    \KwRet{optimized $expression$}\;
}

\Fn{\FSymbolAugmentationStrategy{$node, reward$}}{
    Record expression path and $reward$ of $node$\;
    Update symbol enhancement records based on $reward$\;
    Adjust function library based on records\;
}

\end{algorithm}

\end{document}